\documentclass[10pt,twocolumn,letterpaper]{article}

\usepackage{iccv}
\usepackage{times}
\usepackage{epsfig}
\usepackage{graphicx}
\usepackage{amsmath}
\usepackage{amssymb}
\usepackage{algorithm}
\usepackage{algorithmic}
\usepackage{adjustbox}
\usepackage{hhline}
\usepackage{mathtools}

\iccvfinalcopy 

\def\iccvPaperID{3987} 

\ificcvfinal\fi

\begin{document}

\title{Self-Training and Adversarial Background Regularization for Unsupervised Domain Adaptive One-Stage Object Detection}

\author{Seunghyeon Kim\\
KAIST\\
\and
Jaehoon Choi\\
KAIST\\
\and
Taekyung Kim\\
KAIST\\
\and
Changick Kim\\
KAIST\\
\and
{\tt\small \{seunghyeonkim, whdns44, tkkim93, changick\}@kaist.ac.kr}
}

\maketitle
\ificcvfinal\thispagestyle{empty}\fi

\begin{abstract}
   Deep learning-based object detectors have shown remarkable improvements. However, supervised learning-based methods perform poorly when the train data and the test data have different distributions. To address the issue, domain adaptation transfers knowledge from the label-sufficient domain (source domain) to the label-scarce domain (target domain). Self-training is one of the powerful ways to achieve domain adaptation since it helps class-wise domain adaptation. Unfortunately, a naive approach that utilizes pseudo-labels as ground-truth degenerates the performance due to incorrect pseudo-labels. In this paper, we introduce a weak self-training (WST) method and adversarial background score regularization (BSR) for domain adaptive one-stage object detection. WST diminishes the adverse effects of inaccurate pseudo-labels to stabilize the learning procedure. BSR helps the network extract discriminative features for target backgrounds to reduce the domain shift. Two components are complementary to each other as BSR enhances discrimination between foregrounds and backgrounds, whereas WST strengthen class-wise discrimination. Experimental results show that our approach effectively improves the performance of the one-stage object detection in unsupervised domain adaptation setting.
\end{abstract}



\section{Introduction}
Object detection is a fundamental and core problem in computer vision. Recent studies \cite{FRCNN, SSD, YOLO} have achieved remarkable improvements with the advances of deep neural networks and large-scale benchmarks \cite{ImageNet, COCO, VOC}. Deep learning-based object detectors can be categorized as two-stage detectors or one-stage detectors. Two-stage detectors first extract object regions and then refine them through classification and regression \cite{FRCNN, RFCN}. On the other hand, one-stage object detectors \cite{YOLO, SSD, RefineDet, RFBNet, STDN} directly estimate the coordinates and classes of objects without the Region of Interest (RoI) pooling procedure.

\begin{figure} [t] 
	\centering
		\includegraphics[width=0.48\textwidth]{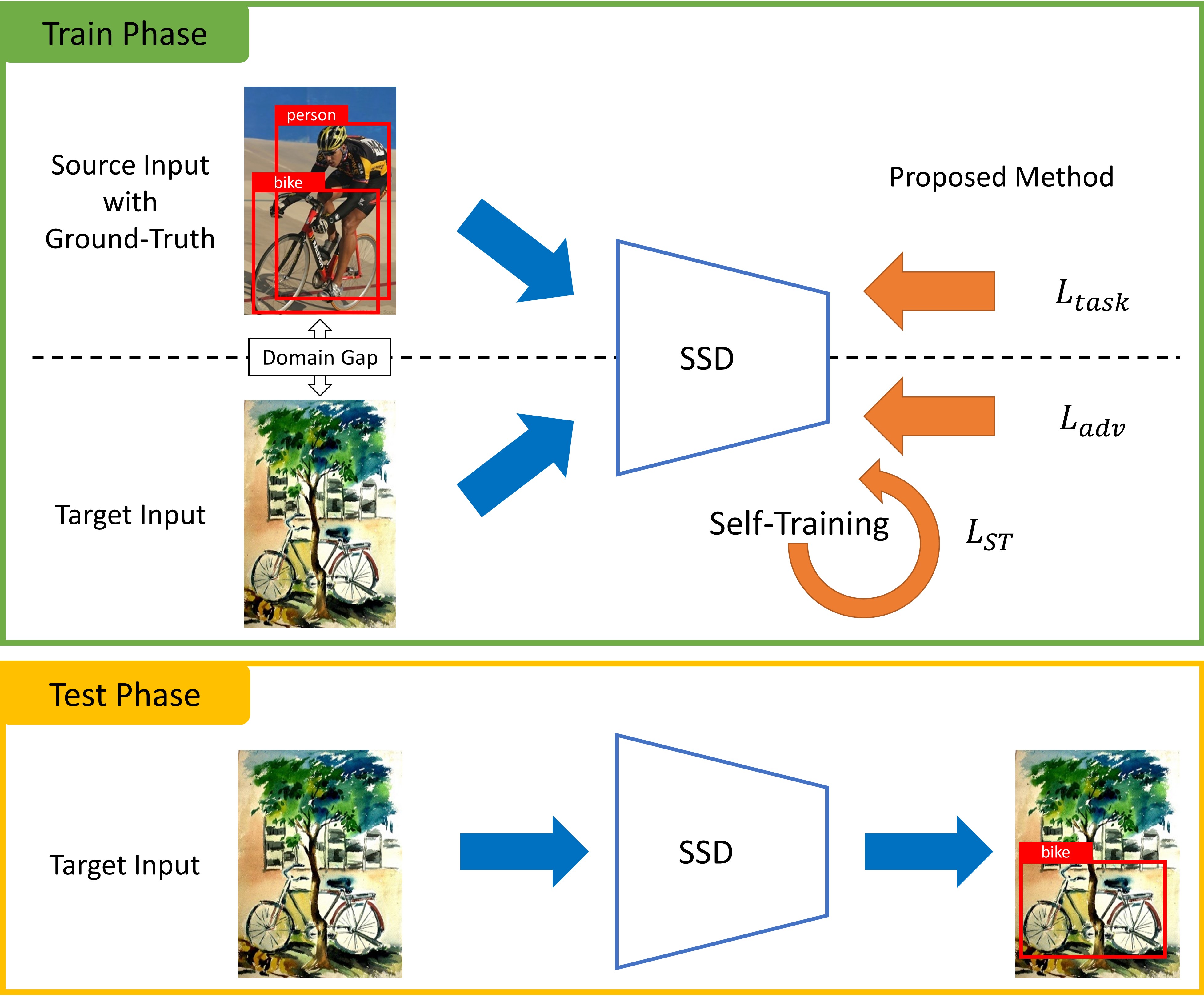}
	\caption{Illustration of unsupervised domain adaptive one-stage object detection. We train an object detector with labeled source images and unlabeled target images. Our method improves the performance of the network for target inputs.}
	\label{fig:intro}
\end{figure}

One limitation of these supervised learning-based methods is the assumption that test data have the same distribution as train data. However, domain shift frequently occurs in many practical applications. For example, variances of object appearance, viewpoints, backgrounds, illumination, and weather condition can degenerate the performance of the network. One possible solution is collecting labeled data for a new domain, but it is usually expensive and time-consuming. To address this issue, domain adaptation transfers knowledge from the train data domain (source domain) to the test data domain (target domain). Especially, unsupervised domain adaptation assumes there are no labels available in the target domain. The goal of domain adaptation is to train the network that performs well on the target domain dataset.



Unfortunately, domain adaptive object detection has received less attention in contrast to classification \cite{MMD, GRL, ADDA, DIRTT, CAN, CDAN} and semantic segmentation \cite{cycada, CGAN, LSD, seg-class-balanced-ST, seg-ProxyLabels}. For object detection, the authors of \cite{DAFRCNN} propose a global feature alignment approach in an adversarial way. 
However, the global feature alignment is not sufficient for object detection since it will align non-transferable backgrounds.
Besides, this work is designed only for a two-stage object detector, Faster R-CNN \cite{FRCNN}. On the other hand, a recent work \cite{Cross-Domain} presents cross-domain weakly-supervised object detection on SSD \cite{SSD}, which is a representative one-stage object detector. However, this work assumes that image-level labels are available on the target domain. Unlike prior works, we introduce one-stage object detection under unsupervised domain adaptation setting. Figure \ref{fig:intro} summarizes the overall framework of our method.

\begin{figure} [t] 
	\centering
		\includegraphics[width=0.48\textwidth]{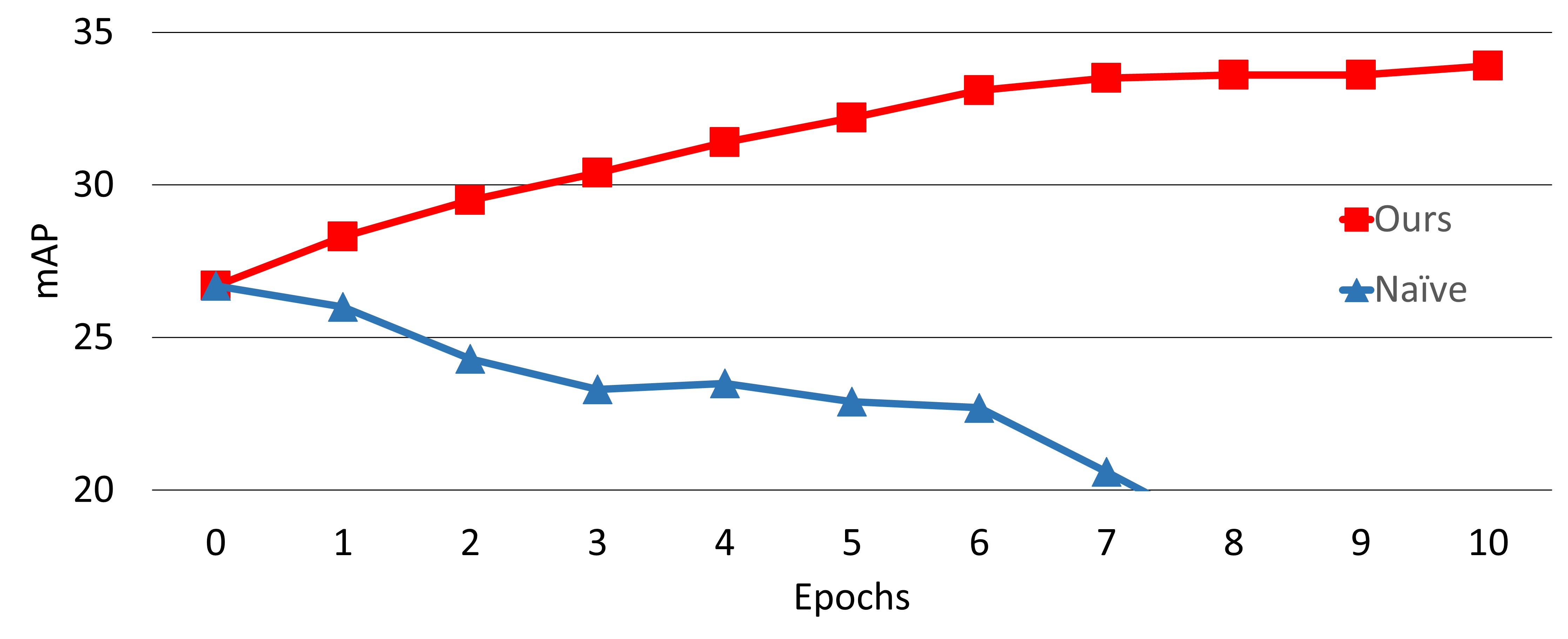}
	\caption{Trends of mAP on the target domain with training epochs. A naive self-training degenerates the accuracy without image-level labels and the regression loss (blue, triangle). Our weak self-training (WST) enables effective self-training under the same settings (red, rectangle).}
	\label{fig:effective}
\end{figure}

In this paper, we propose weak self-training (WST) for stable learning procedure and adversarial background score regularization (BSR) to reduce domain shifts. Previous studies \cite{DIRTT, CAN, Self-Ensembling, seg-class-balanced-ST, seg-ProxyLabels, Cross-Domain} show the effectiveness of self-training for domain adaptation. However, naive self-training approaches without image-level labels harmful for object detectors as shown in Fig. \ref{fig:effective}. To achieve robust self-training, WST minimizes adverse effects of both false positives and false negatives occurred in pseudo-labels. WST does not require any label on the target domain contrary to weakly-supervised approaches \cite{zigzag, self-paced-detection, weakly-mining-pseudo, Cross-Domain} that utilize image-level labels for a reliable choice of pseudo-labels.
Besides, we add BSR at the training phase to reduce the domain shift. We point out that backgrounds of the source and the target data have less common features than foregrounds. From this motivation, BSR extracts discriminative features for target backgrounds. This objective is crucial for one-stage detectors since they do not have region proposal process. 
WST and BSR are complementary to each other as BSR considers discrimination between foregrounds and backgrounds, and WST provides category information of detections.

The contribution of our paper is as follows.
\begin{itemize}
  \item We introduce weak self-training (WST) for domain adaptive object detection which reduces negative effects of inaccurate pseudo-labels.
  \item We propose adversarial background score regularization (BSR) to reduce the domain shift by extracting discriminative features for target backgrounds.
  \item Experimental results show that our approach improves the performance of one-stage object detection under unsupervised domain adaptation setting.
\end{itemize}

\begin{figure*} [t] 
	\centering
		\includegraphics[width=\textwidth]{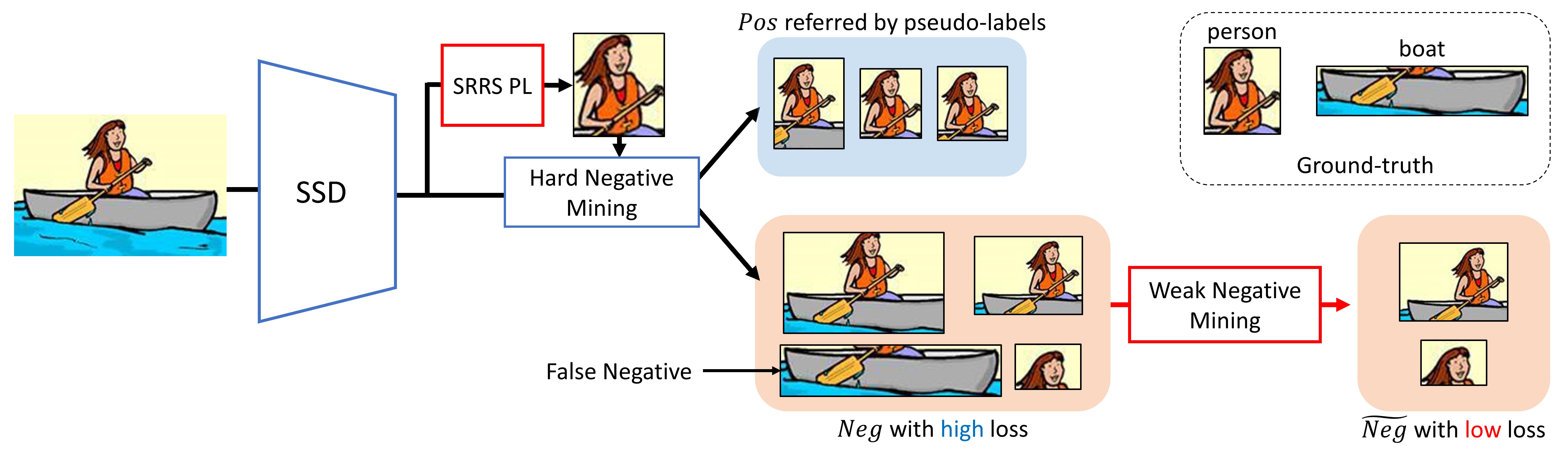}
	\caption{The framework of proposed weak self-training. First, we generate pseudo-labels using SRRS (Supporting Region-based Reliable Score) as a criterion. Second, by following traditional hard negative mining, we obtain two sets of positive examples and negative examples ($Pos$ and $Neg$ respectively). Finally, since the chosen hard negatives are risky for self-training, we select easy samples among $Neg$ and construct $\widetilde{Neg}$. We use examples in $Pos$ and $\widetilde{Neg}$ for weak self-training.}
	\label{fig:ST}
\end{figure*}
\section{Related Work}
\subsection{Object Detection}
Two-stage detectors first extract foreground proposals and then refine the results in the second stage. R-CNN \cite{RCNN} utilizes selective search for region proposal and convolutional neural network for classification. For a faster inference, Fast R-CNN \cite{fast-rcnn} shares the feature map from the same input images. Faster R-CNN \cite{FRCNN} further improves the performance via replacing selective search with a fully convolutional network called region proposal network (RPN).

On the other hand, one-stage detectors also have been researched and show impressive performance on the inference speed. YOLO \cite{YOLO} is a fast object detector based on a fully convolutional network. SSD \cite{SSD} improves the performance of accuracy by utilizing feature maps from various scales. The authors of \cite{focal} point out that one-stage detectors suffer from the class imbalance problem between foregrounds and backgrounds, and propose focal loss which focuses on hard examples rather than easy ones. Furthermore, recent studies \cite{RefineDet, RFBNet, STDN} have improved the performance both in accuracy and inference speed maintaining the efficiency of one-stage detectors.

\subsection{Domain Adaptation}
Domain adaptation reduces the domain gap between train data and test data.
For classification, most recent methods manage to reduce the discrepancy between feature distributions of the source and the target. Early work \cite{MMD} uses Maximum Mean Discrepancy (MMD) as a metric for the discrepancy between two distributions. The authors of \cite{GRL} propose adversarial learning by adding a gradient reversal layer (GRL) and a discriminator to extract domain invariant features. Recent works \cite{ADDA, CDAN, DIRTT, CAN} are based on the domain adversarial learning framework and further improve the discriminative property on the target domain.

For semantic segmentation, GAN-based domain adaptation approaches are shown to be effective. The authors of \cite{cycada} point out that earlier feature-level distribution matching approaches fail to capture pixel-level domain shifts. They propose adversarial domain adaptation model, which aligns both pixel-level and feature-level distribution. In \cite{CGAN}, conditional GAN was used to model the residual of the feature map between the source and target domain.

Some prior works use self-training \cite{SSL, PL} to compensate for the lack of categorical information for either classification \cite{co_training_DA, JDA, tri, semantic_representation_UDA, Self-Ensembling, CAN, DIRTT, PL_curriculum} or segmentation \cite{seg-class-balanced-ST, seg-ProxyLabels}. 

\subsection{Domain Adaptive Object Detection}
Compared to classification and semantic segmentation, domain adaptive object detection has received less attention. The authors of \cite{DAFRCNN} present two domain adaptation components, image-level adaptation and instance-level adaptation. They adopt domain adversarial approach using a discriminator for each component. Recently, the focal loss is utilized for a weak global alignment \cite{strong-weak}. However, it is hard to apply their algorithm to one-stage detector since the algorithms are designed for Faster R-CNN \cite{FRCNN}. The authors of \cite{diversify_and_match} utilize style transfer method, but the method needs large amount of augmented data.

For one-stage object detection, the authors of \cite{Cross-Domain} propose pseudo-labeling and pixel-level adaptation on a cross-domain weakly-supervised setting, which assumes that image-level labels are available for all target images. They first fine-tune the detector with style transferred source images. After that, they generate pseudo-labels by simply choosing the top-1 confidence detections considering image-level labels. They further improve the performance via fine-tuning the network with generated pseudo-labels. However, we confirmed that this method is not valid with the unsupervised domain adaptation setting and degenerates the detection performance. 


On the other hand, proposed WST achieves stable learning without image-level labels and BSR extracts discriminative features for target backgrounds instead of aligning non-transferable features. 



\section{Proposed Method}
In this section, we introduce details of WST and BSR. We adopt SSD \cite{SSD} as our baseline. 

\subsection{Problem Setting}
We assume that source data $(x^s, y^s)$ is drawn from the source domain $X_s$, and target data $(x^t, y^t)$ is drawn from the target domain $X_t$. Here, $x$ is an image and $y=(b, c)$ is a corresponding label, where $b$ is the coordinates of the bounding box and $c$ is the class to which the object belongs. We denote the distribution of domain $X$ as $P(X)$, and $P(X_s) \neq P(X_t)$. We assume that both source data and target data have $K+1$ classes including the background. We set $c$ to 0 for the background. We do not have access to target labels, $y^t$.

We denote layers before $conv5$ of SSD by $F$, and the others by $C$. The output of the SSD is $O=\{r_i\}_{i=1}^n$, where $r_i$ is the $i^{th}$ detection and $n$ is the total number of detections (e.g., $n=8732$ for SSD300). We only take remaining detections after Non-Maximum Suppression (NMS) as final detections. We denote the final outputs as $O^*=\{r_l^*\}_{l=1}^{n^*}$, where $r_l^*$ is the $l^{th}$ detection and $n^*$ is the number of detections in the final outputs.

\subsection{Weak Self-Training}
We propose a weak self-training scheme (WST) to compensate for the lack of categorical information on the target domain. 
Unfortunately, the base network often produces incorrect outputs with high confidence due to large domain shift.
These misclassified outputs become false positives when we choose them for pseudo-labels. Also, false negative error occurs when the network fails to detect some objects in an image. To overcome such problems, WST is designed to omit unreliable examples from the training procedure. The framework of WST is shown in Fig. \ref{fig:ST}.




\bigbreak\noindent
\textbf{Reducing False Negatives.} We minimize the effects of false negatives by modifying the training loss for supervised learning. As defined in \cite{SSD}, the original loss is
\begin{equation} \label{SupervisedLoss}
\begin{aligned}
    L_{task}(x^t, \hat{b}, \hat{c}) = & -\sum_{i \in Pos}{\log(p_i(\hat{c}_i|x^t))} - \sum_{i \in Neg}{\log(p_i(0|x^t))} \\
    & + L_{loc}(x^t, \hat{b}),
\end{aligned}
\end{equation}
where $Pos$ and $Neg$ are sets of positive and negative examples respectively, $\hat{b}$ is a pseudo-bounding box label, $\hat{c}_i$ is a pseudo-class label of $i^{th}$ detection, $p_i(\hat{c}_i|x^t)$ and $p_i(0|x^t)$ are the probability values of the class $\hat{c}_i$ and the background of $i^{th}$ detection, and $L_{loc}$ is the localization loss. However, we observe that Eq. \eqref{SupervisedLoss} is not effective for self-training. Especially, false negatives selected by hard negative mining are harmful to training. We reduce the false negatives by masking out the gradients of background examples during the training. However, it is undesirable to neglect all the background examples because the network will be biased to the foregrounds. Thus, we ignore background examples that have the potential of being foregrounds. 

Negative examples in the $Neg$ set have a large potential of being foregrounds since hard negative mining refers to incorrect pseudo-foregrounds and selects background examples with the highest confidence loss values. For example, in Fig. \ref{fig:ST}, conventional hard negative mining will select the false negative example of the boat as a background example. Thus, we choose $||Neg||/3$ examples that have the lowest confidence loss value among negative examples in $Neg$. We call this process as weak negative mining, and the obtained set from the process is denoted as $\widetilde{Neg}$. Additionally, we do not update the network for bounding box regression since pseudo-labels usually have inaccurate bounding box information. Finally, the modified loss function for weak self-training is defined as

\begin{equation} \label{Weak-Self-Trainig}
\begin{aligned}
    L_{ST}(x^t, \hat{c}) = -\sum_{i \in Pos}{log(p_i(\hat{c}_i|x^t))} - \sum_{i \in \widetilde{Neg}}{log(p_i(0|x^t))}.
\end{aligned}
\end{equation}



\begin{figure*} [t] 
	\centering
		\includegraphics[width=\textwidth]{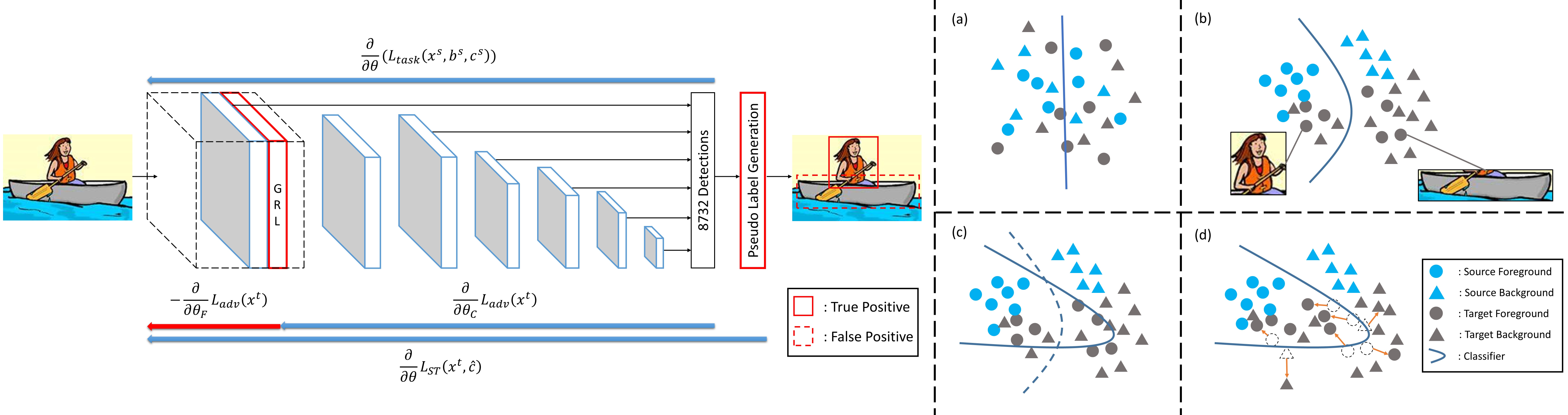}
	\caption{Left: The network architecture with the training losses. $\theta$ represents the parameters of SSD. We add Gradient Reversal Layer (GRL) after $relu4\_3$ of SSD300. Note that GRL is activated only when the target input $x^t$ is used for BSR ($L_{adv}(x^t)$). The example input image is from the target domain. Right: We present feature space representations of the adversarial learning. Each point is a detection. $(a)$: Initial state. $(b)$: The network is trained with the source data. Source examples are well classified, while the boat object of the target data is misclassified as the background. $(c)$: The classifier $C$ minimizes the adversarial loss. The boundary passes through the clusters of the target data. $(d)$: The feature extractor $F$ maximizes the adversarial loss. The example points move away from the boundary.}
	\label{fig:network}
\end{figure*}

\begin{algorithm}[t]
    \caption{Generating Pseudo Labels}
    \label{alg:SRRS}
    \hspace*{\algorithmicindent} \textbf{Input} $O, O^*, \epsilon, \delta$ \\
    \hspace*{\algorithmicindent} \textbf{Output} $\hat{Y}=\{\hat{y}\}$
    \begin{algorithmic}[1]
        \FOR{\texttt{$r^*_l \in O^*$}}
            \FOR{\texttt{$r_i \in O$}}
                \IF{$IoU(r_i, r^*) \geq \delta$}
                    \STATE Collect $r_i$ from $O$
                \ENDIF
            \ENDFOR
            \STATE Calculate $SRRS(r^*_l)$ by Eq. \eqref{SRRS}
            \IF{$SRRS(r^*_l) \geq \epsilon$}
                \STATE Add $r^*_l$ into the set of pseudo-labels, $\hat{Y}$
            \ENDIF
        \ENDFOR
    \end{algorithmic}
\end{algorithm}

\bigbreak\noindent
\textbf{Reducing False Positives.} We propose a criterion for instance-level pseudo labeling based on supporting RoIs. Here, supporting RoIs denote examples having IoU value larger than some threshold $\delta$ with the final detection $r^*$. Rather than using only a single confidence score of a detected box, we consider all the boxes close to the final detection $r^*$. We define Supporting Region-based Reliable Score (SRRS) as
\begin{equation} \label{SRRS}
\begin{aligned}
    SRRS(r^*) = \frac{1}{N_s} \sum_{i=1}^{N_s}{IoU(r_i, r^*) \cdot P(c^*|r_i)},
\end{aligned}
\end{equation}
where $N_s$ is the number of supporting regions, $IoU(A, B)$ denotes the IoU value between region A and region B, $c^*$ is a predicted class of $r^*$, and $P(c^*|r_i)$ is a probability for a region $r_i$ to belong to the $c^*$. By thresholding the score with $\epsilon$, we choose reliable detections. The pipeline of generating pseudo-labels is described in Algorithm \ref{alg:SRRS}.


\subsection{Adversarial Background Score Regularization}
We point out that backgrounds of the source domain and target domain share less common features compared to those of foregrounds. We claim that simple global feature alignment enforces to align non-transferable backgrounds, and makes the training procedure unstable. Motivated by \cite{osb}, we propose background score regularization (BSR) in an adversarial way. The loss function for BSR can be defined as the binary cross entropy function as follows:
\begin{equation} \label{background_score_regularization}
\begin{aligned}
    L_{adv}(x^t) = & -t \sum_i \log(p_i(0|x^t)) \\
    & -(1-t) \sum_i \log(1 - p_i(0|x^t)),
\end{aligned}
\end{equation}
where $i$ is the index of a detection, and $t \in [0, 1]$ is a target value of $p_i(0|x^t)$. As we minimize the loss, the value of $p_i(0|x^t)$ becomes close to $t$. On the contrary, $p_i(0|x^t)$ should be close to 0 or 1 to maximize the loss.

At the training phase, both the classifier $C$ and the feature extractor $F$ minimize the supervised loss $L_{task}$ for the source inputs. For the target inputs, we regularize $C$ to predict the value of $p_i(0|x^t)$ close to $t$ by minimizing $L_{adv}$. On the other hand, $F$ is trained to maximize the adversarial loss to make $p_i(0|x^t)$ close to 0 or 1. Thus, $F$ will learn discriminative features to deceive the classifier. The overall training objectives for BSR can be written as follows:
\begin{equation} \label{overall_F}
\begin{aligned}
    \min_{F} L_{task}(x^s, y^s) - L_{adv}(x^t),
\end{aligned}
\end{equation}

\begin{equation} \label{overall_C}
\begin{aligned}
    \min_{C} L_{task}(x^s, y^s) + L_{adv}(x^t).
\end{aligned}
\end{equation}
We enable the adversarial training using gradient reversal layer (GRL) right after $relu4\_3$ of SSD300.

Without BSR, the base network will produce incorrect prediction with high confidence. The proposed adversarial background score regularization can be thought of as training the classifier to predict with less certainty for target inputs and training the feature extractor to deceive the classifier. In the end, the feature extractor will extract discriminative features, which are easily classified. Figure \ref{fig:network} depicts the process of the adversarial learning.


However, it is not desirable to apply the adversarial loss for all examples because the output of the object detector has an enormous number of background examples. Thus, we sort all examples by their background scores in ascending order and choose low $3N$ examples. Here, $N$ is the number of examples predicted as foregrounds. We found that applying this selection in a batch-wise helps to stabilize the learning. In other words, we combine all examples in a single batch and choose the samples among them. Furthermore, we add a focal term in Eq. \eqref{background_score_regularization} to make the loss numerically stable and still effective. The final adversarial loss is defined as
\begin{equation} \label{background_score_regularization_focal}
\begin{aligned}
    L_{adv}(x^t) = & -t \sum_i |t-p_i(0|x^t)|^{\gamma} \cdot \log(p_i(0|x^t)) \\
    & -(1-t) \sum_i |t-p_i(0|x^t)|^{\gamma} \cdot \log(1 - p_i(0|x^t)),
\end{aligned}
\end{equation}
where $\gamma$ is a hyperparameter. We set $t$ to 0.5 in our experiments. 



Combining BSR and WST will complement each other. BSR reduces domain gaps by helping the network to extract discriminative features for the background class. On the other hand, the network learns category information through WST.

\begin{table*}[t]
  \centering
  \begin{adjustbox}{width=1\textwidth}
  \renewcommand{\arraystretch}{1.2}
  \begin{tabular}{l|cc|ccccccccccccccccccccc}
    Method & BSR & WST & aero & bike & bird & boat & bottle & bus & car & cat & chair & cow & table & dog & horse & mbike & person & plant & sheep & sofa & train & tv & mAP \\ \hline
    Base \cite{SSD} & & & 27.3 & 60.4 & 17.5 & 16.0 & 14.5 & 43.7 & 32.0 & 10.2 & 38.6 & 15.3 & 24.5 & 16.0 & 18.4 & 49.5 & 30.7 & 30.0 & 2.3 & 23.0 & 35.1 & 29.9 & 26.7 \\
    ST & & & 11.8 & 16.4 & 9.1 & 10.8 & 0.3 & 17.7 & 13.9 & 9.1 & 14.7 & 4.5 & 11.1 & 9.1 & 2.3 & 15.2 & 9.1 & 23.7 & 1.8 & 9.1 & 4.5 & 19.8 & 10.7 \\ 
    DANN \cite{GRL} & & & 24.1 & 52.6 & \textbf{27.5} & 18.5 & 20.3 & 59.3 & 37.4 & 3.8 & 35.1 & \textbf{32.6} & 23.9 & 13.8 & 22.5 & 50.9 & 49.9 & 36.3 & \textbf{11.6} & 31.3 & 48.0 & 35.8 & 31.8 \\ \hline
    Ours & \checkmark & & 26.3 & 56.8 & 21.9 & 20.0 & 24.7 & 55.3 & 42.9 & 11.4 & 40.5 & 30.5 & 25.7 & \textbf{17.3} & 23.2 & 66.9 & 50.9 & 35.2 & 11.0 & 33.2 & 47.1 & 38.7 & 34.0 \\
    Ours & & \checkmark & \textbf{30.8} & \textbf{65.5} & 18.7 & \textbf{23.0} & \textbf{24.9} & 57.5 & 40.2 & 10.9 & 38.0 & 25.9 & \textbf{36.0} & 15.6 & 22.6 & 66.8 & 52.1 & 35.3 & 1.0 & \textbf{34.6} & 38.1 & 39.4 & 33.8 \\
    Ours & \checkmark & \checkmark & 28.0 & 64.5 & 23.9 & 19.0 & 21.9 & \textbf{64.3} & \textbf{43.5} & \textbf{16.4} & \textbf{42.2} & 25.9 & 30.5 & 7.9 & \textbf{25.5} & \textbf{67.6} & \textbf{54.5} & \textbf{36.4} & 10.3 & 31.2 & \textbf{57.4} & \textbf{43.5} & \textbf{35.7} \\ 
  \end{tabular}
  \end{adjustbox}
  \caption{Comparison of various methods in terms of mAP. For all methods, the base network is SSD300 \cite{SSD}. Pascal VOC2007 trainval and VOC2012 trainval is used for source dataset and Clipart1k is used for target dataset.
   Descriptions of each method is in Sec. \ref{sec:results_comparison}.}
  \label{tab:comparison_clipart}
\end{table*}

\section{Experiments}
In this section, we present details of the implementation and compare our results with other methods. 

\subsection{Datasets and Evaluation}
In our experiments, we used Pascal VOC2007-trainval and VOC2012-trainval dataset \cite{VOC} as a source domain dataset, and Clipart1k, Watercolor2k, or Comic2k dataset \cite{Cross-Domain} as a target domain dataset.

Pascal VOC \cite{VOC} is a real-world image dataset. It provides both instance-level bounding box annotations and pixel-level annotations. VOC2007-trainval and VOC2012-trainval set have total 16,551 images with 20 distinct categories. Clipart1k \cite{Cross-Domain} is a dataset of graphical images which have a large domain gap with real-world images. It provides 1k images and has the same categories as Pascal VOC. We used all images as a target dataset both for training and evaluation. Watercolor2k and Comic2k \cite{Cross-Domain} are also unrealistic datasets. Each dataset provides 2k of images, 1k for a train set and the other 1k for a test set. Both Watercolor2k and Comic2k have 6 classes which also exist in VOC. We used the train set for training and the test set for evaluation.

For all experiments, we evaluated methods using mean average precision (mAP). We set the confidence threshold to 0.05 and the IoU threshold to 0.5.

\subsection{Implementation Details}
In all experiments, we used SSD300 as a base network. Following the original paper \cite{SSD}, inputs were resized to $300 \times 300$, and we applied all augmentations used in the original paper. SGD was used as an optimizer.

\bigbreak\noindent
\textbf{Base Network.} For the base network, only source dataset was used. We trained the network for 120k iterations with an initial learning rate of $1.0 \times 10^{-3}$, momentum of 0.9, weight decay of $5.0 \times 10^{-4}$. We applied learning rate decay of 0.1 at 80k and 100k, so the final learning rate became $1.0 \times 10^{-5}$. From this setting, the base network shows an accuracy of 77.43\% mAP on the VOC2007 test set.

\bigbreak\noindent
\textbf{Adversarial Background Score Regularization.} Both source and target data were used. Each batch is composed of 32 images, 16 from the source domain, and the other 16 from the target domain. We used $t=0.5$ and $\gamma=2.0$ in Eq. \eqref{background_score_regularization}. We trained the network for 50k iterations with a learning rate of $1.0 \times 10^{-3}$, and reduced the learning rate to $1.0 \times 10^{-4}$ for another 10k.

\begin{table}[t]
  \centering
  \begin{adjustbox}{width=0.48\textwidth}
  \renewcommand{\arraystretch}{1.2}
  \begin{tabular}{l|cc|ccccccc}
    Method & BSR & WST & bike & bird & car & cat & dog & person & mAP \\ \hline
    Base \cite{SSD} & & & 77.5 & 46.1 & 44.6 & 30.0 & 26.0 & 58.6 & 47.1 \\
    ST & & & 78.9 & \textbf{48.1} & 44.9 & 30.1 & 29.1 & 61.7 & 48.8 \\ 
    DANN \cite{GRL} & & & 73.4 & 41.0 & 32.4 & 28.6 & 22.1 & 51.4 & 41.5 \\ \hline
    Ours & \checkmark & & \textbf{82.8} & 43.2 & \textbf{49.8} & 29.6 & 27.6 & 58.4 & 48.6 \\
    Ours & & \checkmark & 77.8 & 48.0 & 45.2 & 30.4 & 29.5 & \textbf{64.2} & 49.2 \\
    Ours & \checkmark & \checkmark & 75.6 & 45.8 & 49.3 & \textbf{34.1} & \textbf{30.3} & 64.1 & \textbf{49.9} \\
  \end{tabular}
  \end{adjustbox}
  \caption{Comparisons on Watercolor2k test set.}
  \label{tab:Watercolor}
\end{table}

\bigbreak\noindent
\textbf{Weak Self-Training.} We fine-tuned the base model with generated pseudo-labels. This method is different from PL in \cite{Cross-Domain} because we update pseudo-labels for every iteration. The network was trained for 10 epochs with a batch size of 32 and a learning rate of $1.0 \times 10^{-5}$. In the process of generating pseudo-labels, we set $\epsilon$ to 0.8 in Algorithm \ref{alg:SRRS} unless otherwise stated. Also, $\delta = 0.5$ was used since the evaluation metric regards an example as a positive when it has a value of IoU larger than 0.5 with the ground-truth.

\bigbreak\noindent
\textbf{BSR with WST.} We followed all the settings of BSR. WST was employed after 50k iterations as the network is not reliable before then. The training was early stopped at 55k iterations since self-training is not helpful when it is overused. The threshold of $\epsilon = \frac{1}{1+e^{-3p}}$ was taken where $p=\frac{current\,iteration}{max\,iteration}$. 

\bigbreak\noindent
\textbf{DANN.} The same configuration with BSR was used to implement DANN for SSD. We aligned distribution of features extracted from $relu4\_3$.

\subsection{Results and Comparisons} \label{sec:results_comparison}
We compared our method with the base network \cite{SSD}, DANN \cite{GRL} and ST. Here, ST is the naive approach of self-training that utilizes pseudo-labels as ground-truth without localization loss. While PL in \cite{Cross-Domain} generates pseudo-labels only once before training, ST and our method recreate pseudo-labels for every single iteration. By comparing DANN and ST with our method, we can directly confirm the effectiveness of the proposed algorithm.

\begin{table}[t]
  \centering
  \begin{adjustbox}{width=0.48\textwidth}
  \renewcommand{\arraystretch}{1.2}
  \begin{tabular}{l|cc|ccccccc}
    Method & BSR & WST & bike & bird & car & cat & dog & person & mAP \\ \hline
    Base \cite{SSD} & & & 43.3 & 9.4 & 23.6 & 9.8 & 10.9 & 34.2 & 21.9 \\
    ST & & & 27.3 & 9.1 & 17.3 & 1.5 & 9.1 & 20.8 & 14.2 \\ 
    DANN \cite{GRL} & & & 33.3 & 11.3 & 19.7 & \textbf{13.4} & \textbf{19.6} & 37.4 & 22.5 \\ \hline
    Ours & \checkmark & & 45.2 & \textbf{15.8} & 26.3 & 9.9 & 15.8 & 39.7 & 25.5 \\
    Ours & & \checkmark & 45.7 & 9.3 & 30.4 & 9.1 & 10.9 & \textbf{46.9} & 25.4 \\
    Ours & \checkmark & \checkmark & \textbf{50.6} & 13.6 & \textbf{31.0} & 7.5 & 16.4 & 41.4 & \textbf{26.8} \\
  \end{tabular}
  \end{adjustbox}
  \caption{Comparisons on Comic2k test set.}
  \label{tab:Comic}
\end{table}

\bigbreak\noindent
\textbf{Results on Clipart1k.} As shown in Table \ref{tab:comparison_clipart}, our weak self-training method can improve the accuracy of the object detector without any labels in the target domain. On the other hand, the naive approach (ST) degenerates the performance due to the effects of false positives and false negatives occurred in generated pseudo-labels. To validate each component of weak self-training, we did ablation study in Sec. \ref{sec:ablation-ST}. Applying BSR shows performance gaps of 8\% and 2.2\% mAP compared to the baseline and DANN respectively. Without any additional networks such as discriminator, adversarial background score regularization effectively improves the performance. Using both BSR and WST further enhances the performance as they complement each other.

Self-training approaches have inferior performance on the class of sheep, because of the poor performance of the base network, while the domain adaptation methods (DANN and BSR) show improvement of nearly 9\% AP. 




\bigbreak\noindent
\textbf{Results on Watercolor2k and Comic2k.} Comparison of performances on VOC $\rightarrow$ Watercolor2k and VOC $\rightarrow$ Comic2k is shown in Table \ref{tab:Watercolor} and \ref{tab:Comic}. 

In the case of Watercolor2k, we set the learning rate to $1.0 \times 10^{-6}$ for self-training methods, since most of the images contain single instance and thus the network is easily overfitted \cite{Cross-Domain}. Furthermore, images in Watercolor2k have no hard backgrounds such as obstacles. From these reasons, our algorithms show less improvement compared to Clipart1k and Comic2k.

For Comic2k, $\gamma=3.0$ was used in Eq. \eqref{background_score_regularization_focal}. Proposed methods improve the accuracy about 5\% mAP from the base, while DANN method seems no improvement.

\section{Analysis}
We conducted ablation studies on WST and parameter sensitivity experiments on BSR. All experiments in this section use Clipart1k as a target dataset.

\begin{table}[t]
  \centering
  \begin{tabular}{c|ccc|c}
    Method & SRRS & Mask & Weak Mask & mAP \\ \hline
    ST (A) & & & & 10.7 \\
    SRRS (B) & \checkmark & & & 10.5 \\
    Mask (C) &  & \checkmark & & 16.8 \\
    SRRS+Mask (D) & \checkmark & \checkmark & & 29.2 \\
    Weak Mask (E) & & & \checkmark & 31.3 \\
    SRRS+Weak Mask (F) & \checkmark & & \checkmark & 33.8
  \end{tabular}
  \caption{Ablation study on WST. Mask denotes that no negative example is used for learning. Weak Mask indicates that weak negative mining is used for sampling negative examples. Method A is identical to ST in Table \ref{tab:comparison_clipart} and method F is the proposed weak self-training.}
  \label{tab:ST_Ablation}
\end{table}

\begin{figure} 
	\centering
		\includegraphics[width=0.48\textwidth]{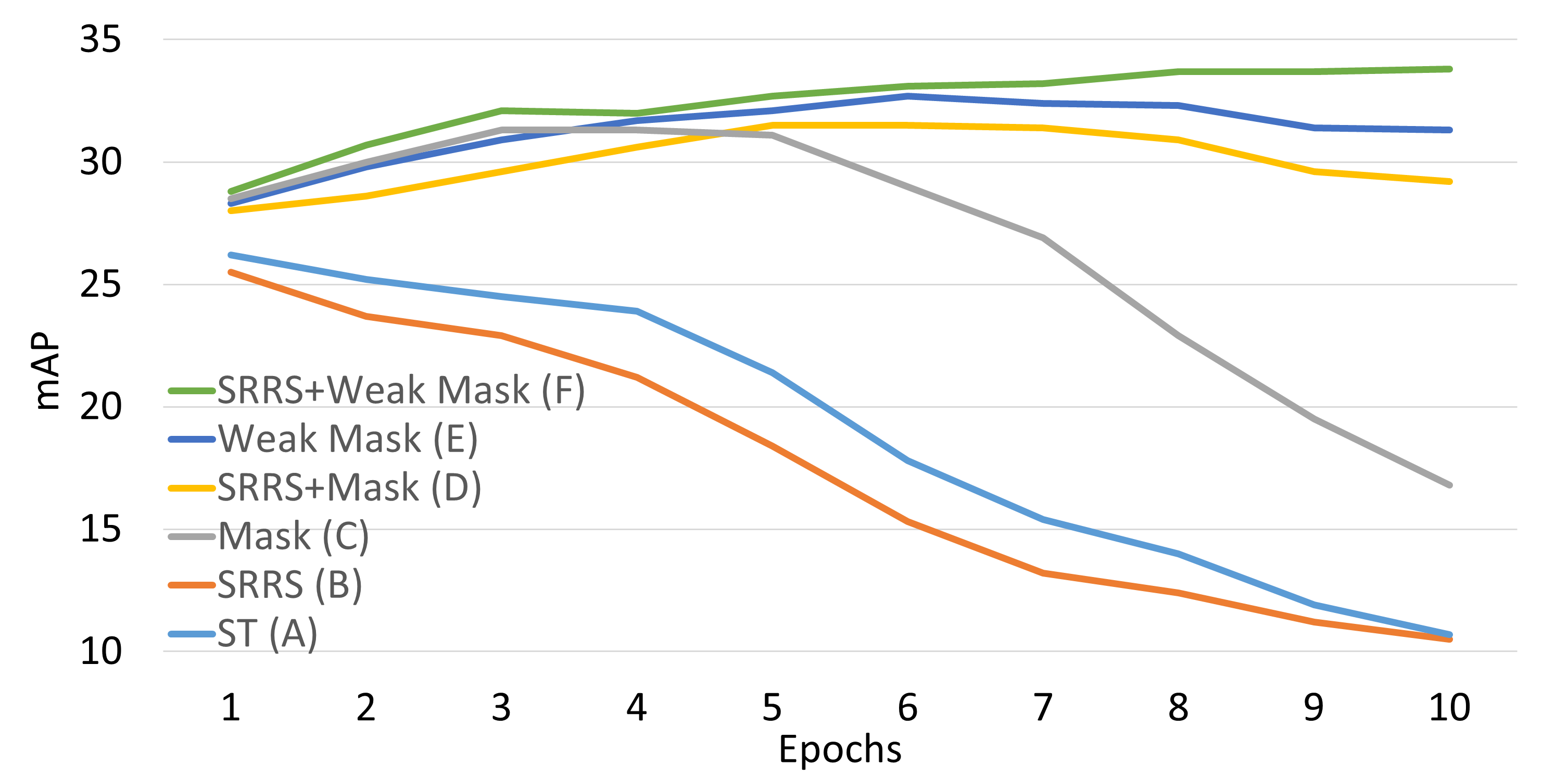}
	\caption{The change of accuracy with the training procedure. We provide all performances during training for each epoch. Different combinations of each method are shown in Table \ref{tab:ST_Ablation}. The performances of method A, B and C decreased dramatically due to the adverse effects of false positives and false negatives in pseudo-labels. The proposed method F had the highest value with the stable learning process.}
	\label{fig:ST-Ablation}
\end{figure}

\subsection{Ablation Study on WST} \label{sec:ablation-ST}
To validate each component of proposed weak self-training, we provided ablation study. In these experiments, we set $\epsilon$ to 0.9 without SRRS, and 0.8 with SRRS in Algorithm \ref{alg:SRRS}. In Table \ref{tab:ST_Ablation}, we present several combinations of three components with their method names and performances. Method A is the naive self-training and method F is the proposed weak self-training. In Fig. \ref{fig:ST-Ablation}, we provide the performance trends with training epochs to validate learning stability.


\begin{table}[t]
  \centering
  \begin{tabular}{c|c|c||c|c|c}
    $t$ & $\gamma$ & mAP & $\gamma$ & $t$ & mAP \\ \hline
     & 0.0 & not converge & & 0.25 & 28.5 \\
     & 1.0 & not converge & & 0.33 & 19.8 \\
    0.5 & 2.0 & \textbf{34.0} & 2.0 & 0.5 & \textbf{34.0} \\
     & 4.0 & 32.9 & & 0.67 & 21.2 \\
     & 5.0 & 31.1 & & 0.75 & 20.8
  \end{tabular}
  \caption{Parameter sensitivity on BSR. The table shows the performance of BSR with various values of $\gamma$ and $t$ in Eq. \eqref{background_score_regularization_focal}.}
  \label{tab:BSR_parameter_sensitivity}
\end{table}

\begin{figure}[t]
	\centering
		\includegraphics[width=0.49\textwidth]{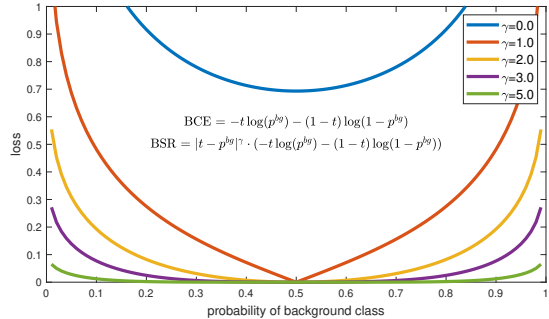}
	\caption{Visualization of background score regularization (BSR) with different values of $\gamma$. $p^{bg}$ denotes the probability of background class. BCE stands for binary cross entropy loss, and it is identical to BSR with $\gamma=0$. Note that $t=0.5$ for all graphs.}
	\label{fig:BSR_focal}
\end{figure}

From the results of method A, B, and C, self-training cannot be accomplished without either SRRS or Mask. The accuracy of the three methods rapidly decreases as training proceeds. This implies that reducing both false positives and false negatives is crucial for object detection.

\begin{figure*}[t]
	\centering
		\includegraphics[width=0.95\textwidth]{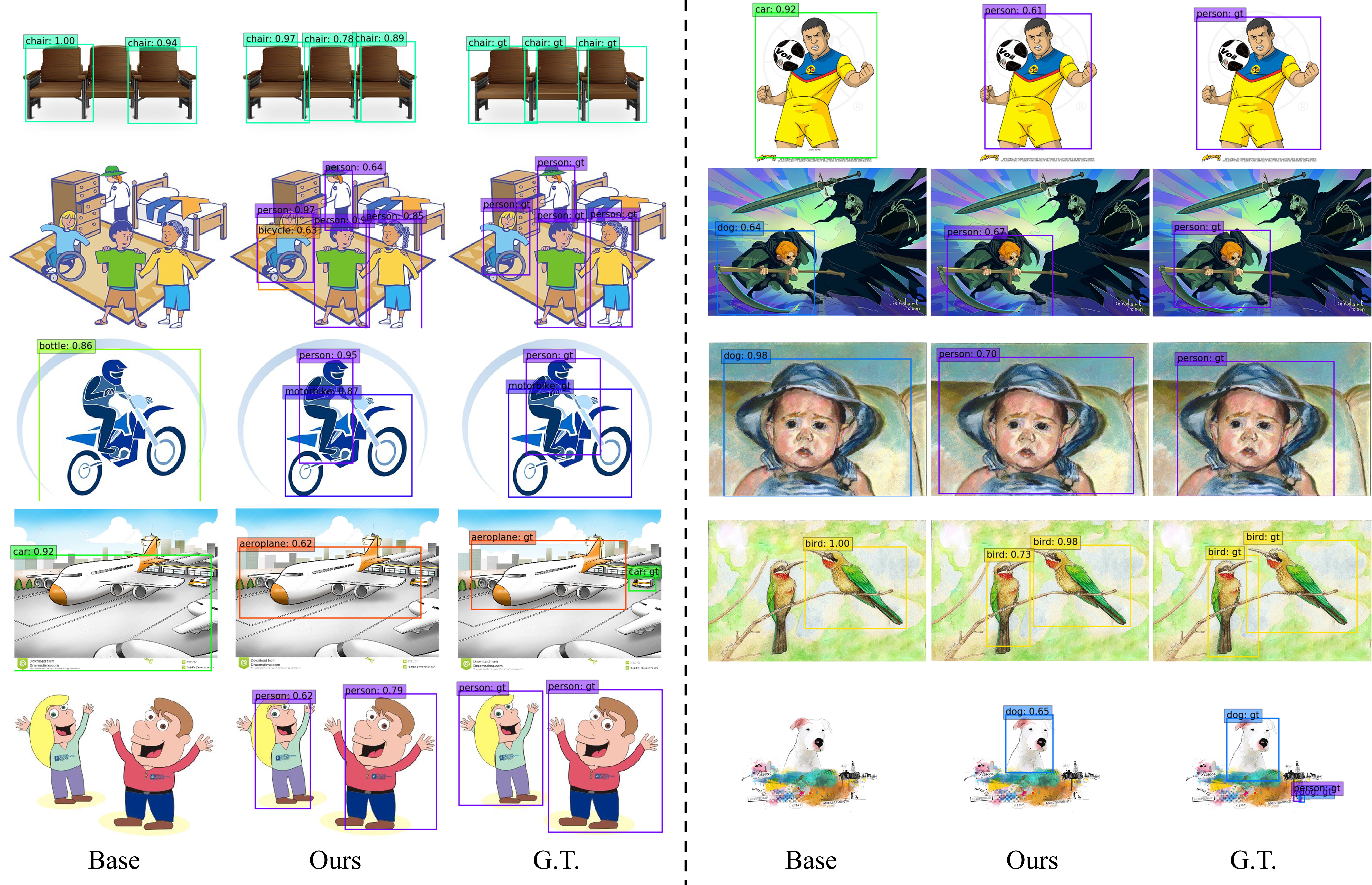}
	\caption{Qualitative results on Clipart1k, Watercolor2k, and Comic2k. We present the results of the base network, our method, and ground-truth from left to right.}
	\label{fig:qualitative}
\end{figure*}

Comparing with Mask, proposed Weak Mask showed remarkable performance improvement by 14.5\% mAP without SRRS and 4.6\% mAP with SRRS (E compared to C and F compared to D). Comparing methods A and E, only adding weak negative mining can improve the performance dramatically. From these results, we confirm that selecting reliable negative samples is even more effective than selecting positive samples. More specifically, since pseudo-labels commonly omit some foreground instances, hard negative mining tends to select those missing instances as backgrounds. Masking all the gradients of the negative examples can stabilize the learning, but the network will be biased to foregrounds as it never learns about backgrounds. Thus, the proposed weak negative mining is critical for a self-training under domain adaptation setting.

We observed that SRRS is not valuable when it is used alone. Combined with either Mask or Weak Mask, SRRS succeeded to enhance both learning stability and accuracy (D compared to C and F compared to E). Although SRRS can reduce not only false positives but also true positives, we experimentally confirmed that reducing false positives is crucial even though less true positives are used for training.



\subsection{Parameter Sensitivity on BSR}
Table \ref{tab:BSR_parameter_sensitivity} shows the results for parameter sensitivity of both $\gamma$ and $t$ in Eq. \eqref{background_score_regularization_focal}. The focusing parameter $\gamma$ controls the strength of the loss. As $\gamma$ gets smaller, more detections contribute to the adversarial training. More specifically, the network will have too strong regularization on backgrounds with $\gamma=0$ and $\gamma=1$. On the other hand, the regularization is relaxed with large values of $\gamma$. The network will ignore examples which have the probability of the background around $0.5$ with large $\gamma$. See Fig. \ref{fig:BSR_focal} for visualization of BSR with various values of $\gamma$. The performance is insensitive to the value of $\gamma$ unless it is too small.

We conducted experiments on $t \in \{\frac{1}{4}, \frac{1}{3}, \frac{1}{2}, \frac{2}{3}, \frac{3}{4}\}$ to verify trends of the learning according to $t$. The network trained with $t=0.5$ shows better performance than the others. For the other values, the network is easily over-regularized and the performance rapidly drops after the learning rate decay. 

\subsection{Qualitative Results}
We compare the qualitative results of the base network, the proposed method, and ground-truth as shown in Fig. \ref{fig:qualitative}. We found that the proposed method detects objects with less confidence but correctly, compared to the base network due to BSR. As shown in top left example in Fig. \ref{fig:qualitative}, the probabilities of two chairs that are detected by the base network are decreased while the chair between them is only detected by our method. 

\section{Conclusion}
In this paper, we have addressed unsupervised domain adaption for one-stage object detection. We enable self-training for object detection by reducing the adverse effects of inaccurate pseudo-labels. Proposed weak self-training (WST) effectively reduces false negatives and false positives by masking the gradients of hard negative examples and utilizing SRRS as a criterion for pseudo-labeling. We have also present adversarial background score regularization (BSR) to reduce the domain shifts by enhancing discrimination between foregrounds and backgrounds of the target data.

\vspace{5mm} \noindent
\textbf{Acknowledgements} This work was supported by Institute for Information \& communications Technology Planning \& Evaluation (IITP) grant funded by the Korea government (MSIP) (No. 2018-0-00198), Object information extraction and real-to-virtual mapping based AR technology)

{\small
\bibliographystyle{ieee_fullname}
\bibliography{egbib}
}

\end{document}